\documentclass{IOS-Book-Article}

\usepackage{mathptmx}
\usepackage{soul}\setuldepth{article}
\usepackage{graphicx}
\usepackage{booktabs} 
\usepackage{multirow}
\usepackage[table]{xcolor}

\def\hb{\hbox to 11.5 cm{}}

\begin{document}

\pagestyle{headings}
\def\thepage{}
\begin{frontmatter}              

\title{Toward an Intelligent Tutoring System for Argument Mining in Legal Texts}


\author[A]{\fnms{Hannes} \snm{Westermann}
\thanks{Corresponding Author: Hannes Westermann, E-mail: hannes.westermann@umontreal.ca}},
\author[B]{\fnms{Jarom\'{i}r} \snm{\v{S}avelka}},
\author[C]{\fnms{Vern R.} \snm{Walker}},
\author[D]{\fnms{Kevin D.} \snm{Ashley}} and
\author[A]{\fnms{Karim} \snm{Benyekhlef}}

\runningauthor{H. Westermann et al.}
\address[A]{Cyberjustice Laboratory, Facult\'e de droit, Universit\'e de Montréal}
\address[B]{School of Computer Science, Carnegie Mellon University}
\address[C]{LLT Lab, Maurice A. Deane School of Law, Hofstra University}
\address[D]{School of Computing and Information, University of Pittsburgh}

\begin{abstract}
We propose an adaptive environment (CABINET) to support caselaw analysis (identifying key argument elements) based on a novel cognitive computing framework that carefully matches various machine learning (ML) capabilities to the proficiency of a user. CABINET supports law students in their learning as well as professionals in their work. The results of our experiments focused on the feasibility of the proposed framework are promising. We show that the system is capable of identifying a potential error in the analysis with very low false positives rate (2.0-3.5\%), as well as of predicting the key argument element type (e.g., an issue or a holding) with a reasonably high F$_1$-score (0.74).

\end{abstract}

\end{frontmatter}

\section{Introduction}
In this paper we examine the application of cognitive computing \cite{licklider1960} to support both a law student learning how to extract key arguments from a court opinion and a legal expert performing the same. We propose an adaptive environment that evolves from a tutoring system to a production annotation tool, as a user transitions from a learner to an expert. The concept is based on a novel cognitive computing framework where (1) the involvement of machine learning (ML) based components is carefully matched to the proficiency level of a human user; and (2) the involvement respects the limitations of the state-of-the-art of automated argument mining in legal cases. We experimentally confirm feasibility of the key ML components by testing the following two hypotheses: Given a sentence in a case brief, it is possible (H1) to detect if the sentence is placed in an \emph{incorrect} section, and (H2) to predict the \emph{correct} section for the sentence.


\section{Background}
\label{sec:background}

%

Lawyers routinely analyze case decisions (i.e., court opinions) to gain insight into what is a persuasive or binding precedent (typically common law countries) and/or what is the established decision-making practice in a given matter (typically civil law countries). As the list of relevant cases may be long and the opinions might be sizeable, a principled approach to the analysis is necessary to make the task feasible and as efficient/effective as possible. Such an approach requires knowing how to read an opinion, which parts to focus on, and which information to identify as crucial for understanding the case.

In  U.S. law schools, case briefs are widely employed to teach law students how to analyze a case and how to use prior decisions to create new arguments or analyses~\cite{Brostoff2013United}. Writing a case brief involves reading and understanding a case, and identifying  text passages that contain the key aspects of the decision. These are then extracted and arranged in a structured format that often includes the following sections:

\begin{itemize}
    \item \textbf{Facts} - Events and actions relevant to the dispute.
    \item \textbf{Issue} - Main questions (points of contention) the court must address.
    \item \textbf{Holding} - Legal rulings when the law is applied to a particular set of facts.
    \item \textbf{Procedural History} - The treatment the dispute has received from the courts.
    \item \textbf{Reasoning} - The analysis of the court leading to the outcome.
    \item \textbf{Rule} - The official rules the court must adhere to (e.g., statutory provisions).
\end{itemize}



Interestingly, many professors never ask  students to turn in their briefs and, hence, do not provide a learner with much needed feedback. \cite{makdisi2009} However, practice and feedback are essential for learning. When it comes to practice, the research clearly shows that it should be focused and deliberate \cite{ericsson2003}, at the appropriate level of challenge \cite{ericsson2003}, and in sufficient quantity \cite{healy1993}. Such practice should be coordinated with targeted feedback on specific aspects of students' performance in order to promote the greatest learning gains.~\cite{ambrose2010,black1998} Feedback should also be timely, i.e., immediate and frequent \cite{hattie2007}. These elements do not seem present when it comes to learning to brief cases. As a result, while law students tend to start out by dutifully briefing cases, they usually switch to a less detailed approach after a few weeks, focused on color-coding sentences or taking notes in the margins of the case texts. Due to the lack of feedback and practice, it is thus unclear whether the crucial skill of briefing cases has been acquired.


To address the issue we propose CABINET, an intelligent tutoring system that gradually evolves from a platform aimed at learners to a powerful annotation environment to support an expert. 
In a nutshell, CABINET allows a user to select a sentence and assign it to one of the case brief's sections. More importantly, the system provides varying levels of scaffolding (i.e., varying levels of challenge) and timely feedback appropriate for the learner's level of proficiency to maximize the learning outcomes. The tool thus adapts with the user, teaching them how to brief cases at first and later supporting them in briefing and understanding cases more efficiently.

\section{Related Work}
\label{sec:related_work}

Numerous researchers have proposed frameworks where a human and a computer complement each other in performing tasks in the legal domain. For example, human-aided computer cognition framework has been proposed and evaluated in the context of eDiscovery. \cite{hogan2009} Active learning has been explored in various contexts, such as classification of German \cite{waltl2017} or United States \cite{savelka2011} statutory provisions, or relevance assessment in eDiscovery \cite{cormack2015}. The annotation tool proposed in \cite{westermann2020} supports human annotators by enabling them to view similar sentences together. The environment described in \cite{westermann2019} provides statistical insights into a data set assisting a human expert in creating text classification rules. The work presented in this paper is to our best knowledge the first study that explicitly maps multiple ML components to different levels of user's proficiency.

Multiple research studies have explored applications of intelligent tutoring systems in teaching legal argumentation and case analysis skills. These include supporting law students in graphically representing legal arguments \cite{pinkwart2009evaluating}, assessing case relevance and distinguishing cases \cite{aleven2003using}, performing case-based and rule-based reasoning \cite{bittencourtab2007themis}, and selecting applicable legal rules from statutes \cite{routen1992reusing} and precedents \cite{muntjewerff2001evaluating}. An adaptive legal textbook based on knowledge graphs has been proposed in \cite{DBLP:conf/aied/SovranoABV22}. The framework presented in this paper is the first intelligent tutoring system for legal domain that can be adapted to the proficiency level of a user to eventually support a legal professional in the task of analyzing cases by extracting their key arguments.




A key component of the proposed framework is the automatic recognition of key argument elements in case texts. This task has been studied extensively in AI \& Law and it is often referred to as automatic identification of rhetorical roles that sentences play in the text of courts’ opinions. Rhetorical role classification focuses on segmenting cases into functional parts \cite{wyner2010,harasta2019,savelka2018} which can, e.g., improve legal information retrieval and enable legal argument retrieval \cite{grabmair2015,walker2019,walker2017}. Information about a sentence's rhetorical role can also be utilized in summarization \cite{farzindar2004,hachey2006,moens2007,bhattacharya2019,Xu2020}. The roles  often include categories such as Facts, Issue, or Conclusion that are related to the ones used in this work.

A variety of ML/NLP techniques have been employed to predict  sentence role labels. These span from rule-based approaches~\cite{westermann2019} to applying   ML models such as Support Vector Machines \cite{walker2019}. The problem has also been treated as tagging of sequences that consist of multiple sentences instead of simpler single sentence classification. Here, models such as Conditional Random Fields have often been used ~\cite{saravanan2006,savelka2017}. A deep learning system based on Bi-LSTM was shown to perform well in \cite{bhattacharya2019}. Systems based on a multilingual embeddings, Bi-LSTM, or pre-trained language models demonstrated strong transfer learning capabilities in this task \cite{savelka2021,savelka2021b}. The work presented in this paper is the first attempt to use the sentence rhetorical role identification models in intelligent tutoring to support law students in learning how to analyze legal cases.

\section{Proposed Framework}
\label{sec:framework}
\begin{figure}
    \includegraphics[width=1\textwidth]{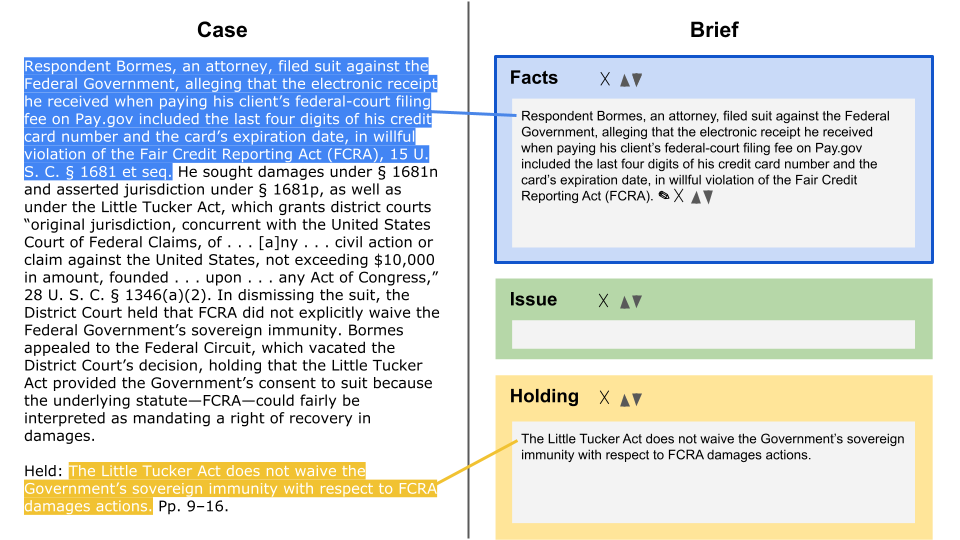}
    \caption{An overall design of CABINET's user interface: The analyzed document is displayed on the left. The case brief sections are populated and edited on the right. The system preserves the mapping between the original text and the resulting case brief sections.}
    \label{fig:cabinet_base}
\end{figure}

We propose CABINET (CAse Brief INteractive EnvironmenT) which is a cognitive computing framework that adapts to the proficiency level of a user. An overall design of CABINET's user interface is shown in Figure \ref{fig:cabinet_base}. We adopt a fairly standard layout where an analyzed document is displayed beside a template to be populated with the extracted key argument elements. The goal is to identify the key argument spans of text (\textit{argument element identification task}) and to categorize them in terms of case brief sections (\textit{argument element categorization task}).

CABINET provides scaffolds and supports that allow it to evolve from an intelligent tutoring system for learners to a tool to support the more efficient work for professionals. 
To this end we adopt the National Institutes of Health's competencies proficiency scale\footnote{\url{https://hr.nih.gov/working-nih/competencies/competencies-proficiency-scale}} (NIH proficiency scale), a highly-regarded instrument used to measure one's ability to demonstrate competency in a task.

\begin{figure}
    \includegraphics[width=1\textwidth]{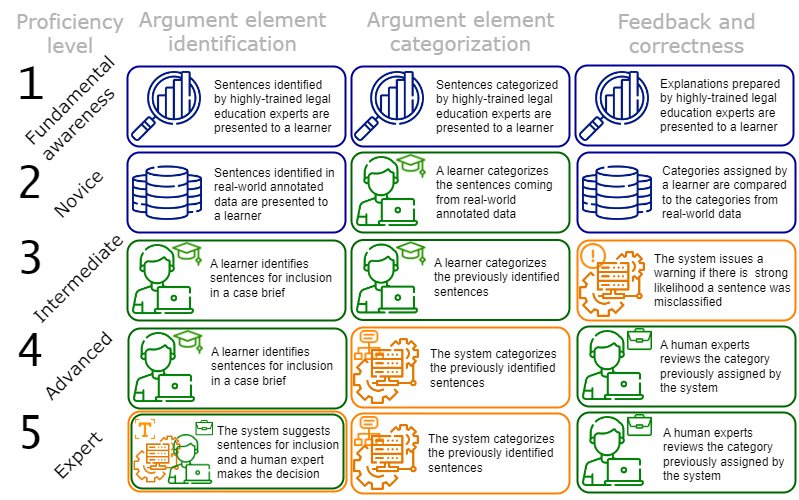}
    \caption{The rows of the diagrams correspond to the five NIH proficiency levels. The columns represent tasks performed as cooperation between a user and a computer. The performance of the tasks is either based on static expert data that have been pre-annotated (blue), user's work (green), or ML component (orange).}
    \label{fig:framework}
\end{figure}

Figure \ref{fig:framework} shows how the system adapts and adjusts to the proficiency level of the user as indicated by the NIH scale, from level 1 (Fundamental Awareness) to level 5 (Expert). Initially, the system provides the user with reference answers and explanations to provide an adequate level of challenge and timely feedback (blue in Figure \ref{fig:framework}). As the user learns, they are able to perform more tasks themselves (green). The system takes on the role of a safe-guard against apparent mistakes relying on one of its ML components. In the latter stages the system's ML components take over the initial steps in performing the work (orange) and a user (now an expert) reviews the results and corrects mistakes. 




\textbf{Fundamental Awareness (NIH level 1)} - 
An individual at this level has common knowledge for understanding basic techniques and concepts. The learner is aware of the concept of briefing a case, its purpose and value, and is superficially aware of case brief sections. At this level, CABINET leverages the so-called \emph{worked example effect}: studying worked examples appears to be more effective than learning by solving the equivalent problems. \cite{sweller2006} This effect has also been  confirmed in the related area of reasoning about legal cases. \cite{nievelstein2013} As shown in Figure \ref{fig:framework}, the learner's task is to use the CABINET interface to inspect and reflect on cases annotated by legal education experts. The interface also provides explanations justifying the choices of the expert. This stage relies on the in-depth annotation of a small number of cases (tentatively 10-20 cases).


\textbf{Novice (NIH Level 2)} - 
At this level, an individual has the level of experience gained in a classroom and/or experimental scenarios or as a trainee on the job. They are expected to need help when performing a skill. Learners at this level are ready to attempt to categorize key argument elements with respect to case brief sections. As shown in the second row of Figure \ref{fig:framework}, learners are presented with texts where the key argument elements have already been identified by legal experts. The learner performs the argument element categorization task. Their choices are compared to those of the experts and the learner is notified about a mismatch and the category assigned by an expert is revealed. 



\textbf{Intermediate (NIH Level 3)} - 
\label{level3}
An individual at this level is able to successfully complete tasks as requested with occasional expert help. At this level, the learner attempts to identify key argument elements in a text and to categorize them. As shown in the third row of Figure \ref{fig:framework}, CABINET assumes the role of a safe-guard which evaluates the work of the advanced learner. Here, the feedback comes from a ML component that identifies the sentences that have a very high likelihood of being placed in the wrong section. The feasibility of such an ML model is evaluated in Section \ref{sec:evaluation} (Experiment 1).



\textbf{Advanced (NIH Level 4)} - 
\label{level4}
At this level, an individual can perform the actions associated with the skill without assistance. It is assumed that a trained professional can independently identify the key argument elements in a text as well as classify them with the correct case brief section. At this stage, CABINET employs a classification model to predict the case brief section of an argument element that a user identified for inclusion in the case brief. The user is expected to evaluate the predictions and correct potential errors. Some errors are tolerable at this stage, since the user is proficient enough to efficiently correct them. 
The feasibility of such a model is evaluated in Section \ref{sec:evaluation} (Experiment 2).

\textbf{Expert (NIH Level 5)} - 
At this level, an individual is  an expert in a given area. They can provide guidance, troubleshoot and answer questions related to this area of expertise and the field where the skill is used. CABINET provides the same kind of assistance as in the previous stage,  but uses more active ways of supporting the professional at this level. Specifically, CABINET subtly highlights passages in a text with colors corresponding to predicted case brief sections and intensity corresponding to system's confidence in the passage being a key argument element. This is achieved by a ML component applied to a full text. Since such predictions cannot be performed with a high degree of reliability (see Section \ref{sec:related_work}), the highlighted text passages are only meant to augment the expert's review of a case text, allowing them to identify the key argument elements more efficiently.

\section{Experiments}
\label{sec:evaluation}
To examine the framework's feasibility,  we assess two hypotheses that correspond to the system's key capabilities described in Section \ref{sec:framework}. Given a sentence in a case brief: 

\begin{enumerate}
    \item[(H1)] \textellipsis\ it is possible to detect if the sentence is in an \emph{incorrect} section.
    \item[(H2)] \textellipsis\ it is possible to predict the \emph{correct} section for the sentence.
\end{enumerate}

\noindent The capability assessed by H1 is deployed at the Intermediate proficiency level (NIH Level 3), to warn a user when a sentence is likely assigned to an incorrect case brief section. 
The capability assessed by H2 is utilized at the Advanced and Expert proficiency levels (NIH Levels 4 and 5), to predict the correct section for a text passage identified by a user as a key argument element.

\subsection{Dataset}
\label{sec:dataset}
\begin{figure}\
    \centering
    \includegraphics[width=0.59\textwidth]{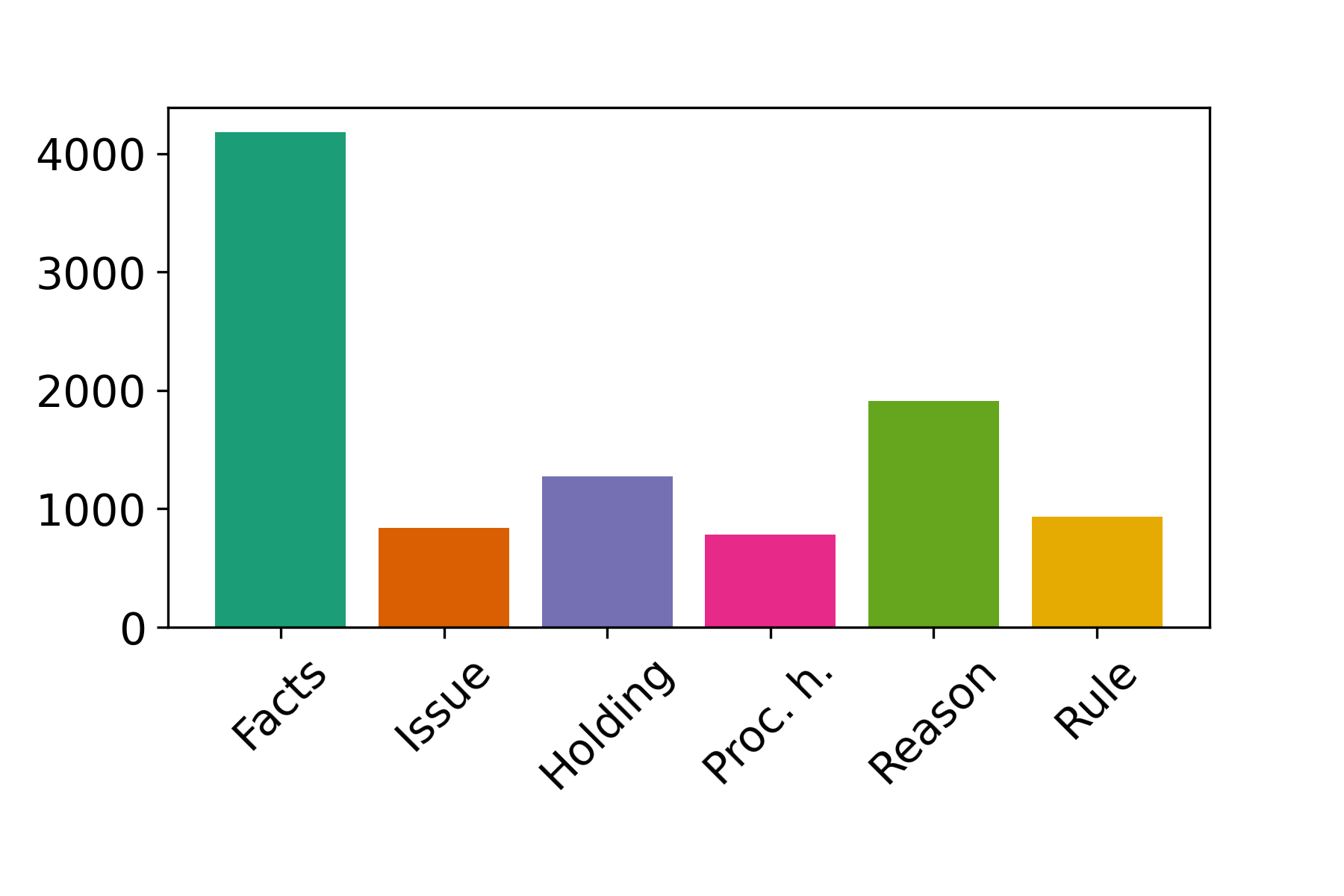}
    \includegraphics[width=0.39\textwidth]{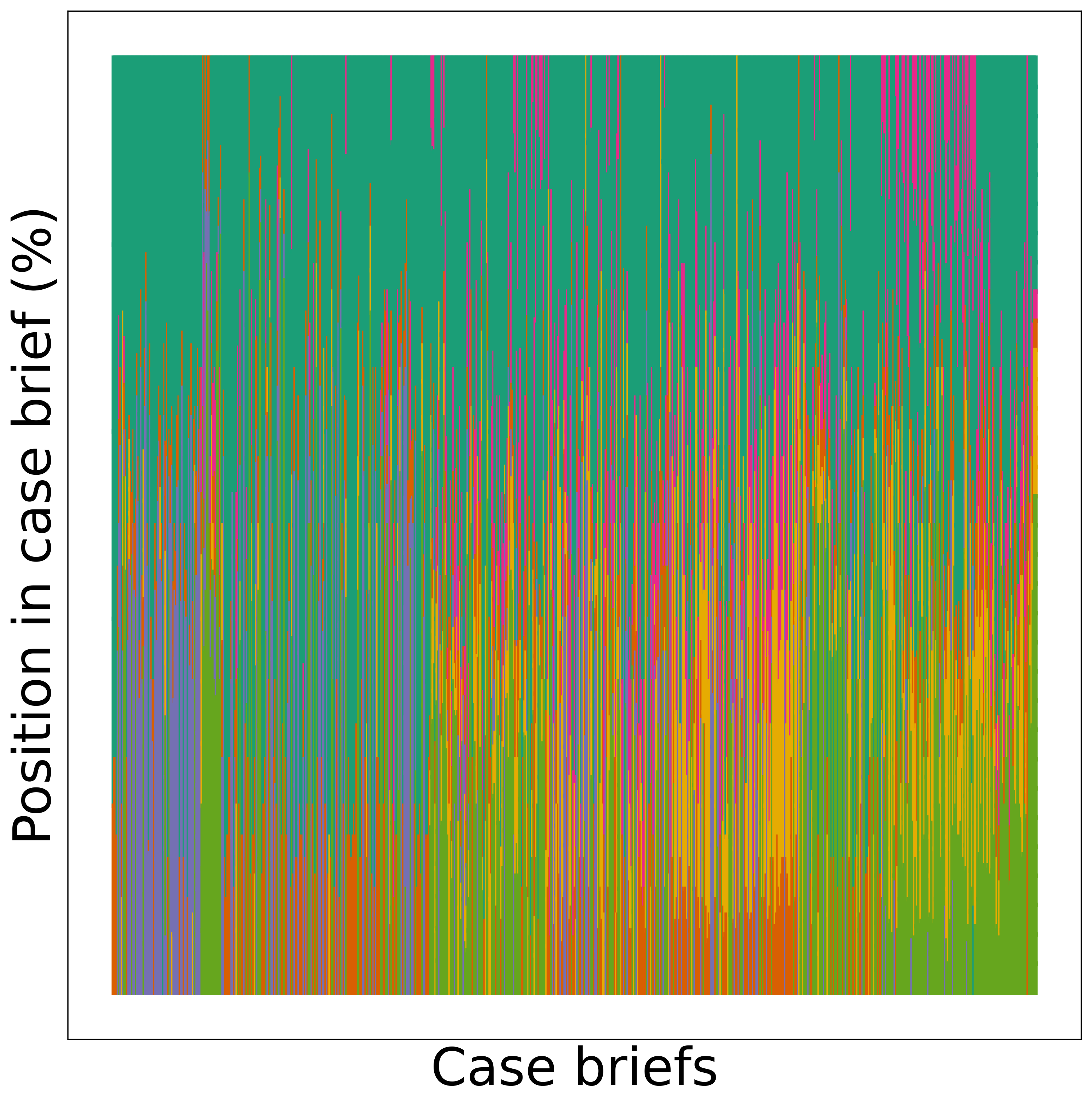}
    \caption{Left: Distribution of sentences in the dataset by section. Right: Location of sentence classes (y-axis) across the different case briefs (x-axis).}
    \label{label_distribution}
\end{figure}

We obtained a dataset of 715 unique case briefs by scraping a publicly available Case Brief Summary database.\footnote{Accessible at: \url{http://www.casebriefsummary.com/}. Currently, the website appears to be off-line. However, it has been archived by the Web Archive project at \url{https://web.archive.org/web/20200927234341/http://www.casebriefsummary.com/}} 
We used an extensive battery of regular expressions to segment the retrieved briefs into individual sections corresponding to the key argument element types. While there were over 100 unique section names we identified the six main types (see Section \ref{sec:background}) to which we could map many of the different variations (e.g., all of ``Legal Issue", ``Issues", and ``Issue" map to a single category). 
We applied a specialized legal case sentence boundary detection system to segment the sections into 9,924 sentences. \cite{savelka2017b} 
Figure \ref{label_distribution} shows the distribution of the sentences in terms of the key argument element types from the perspective of their overall counts as well as their distribution over the individual case briefs.
We divide the dataset into random splits on a document basis. The splits are used for training (70\%), validation (15\%) and testing (15\%).


\subsection{Model and Training}
As a \textbf{baseline} we use a simple model that randomly predicts the labels based on their frequency in the dataset. We use the standard implementation provided by the sklearn framework, with  ``stratified" sampling.\footnote{\url{https://scikit-learn.org/stable/modules/generated/sklearn.dummy.DummyClassifier.html}} As the main model we employ \textbf{RoBERTa} (a robustly optimized BERT pretraining approach) developed by Liu. \cite{liu2019} The model was chosen due to its high performance and simplicity to train. While higher performance might be achieved by more recent models, the RoBERTa model suffices for the purpose of proving the feasibility of the key components of the framework. Out of the available models we chose to work with the smaller \emph{roberta.base} that has 125 million parameters, for faster iteration times.
Due to the sentences being short, we did not have to address the model's sequence length limitation of 512 tokens. The training set is used to train the model for 4 epochs. At each epoch, we evaluate the performance of the model against the evaluation set and pick the best-performing model for our experiments. 

\subsection{Testing H1: Warning about an Incorrectly Categorized Argument Element}
When interacting with an Intermediate user (NIH Level 3) CABINET issues a warning if a piece of text identified by the user as a key argument element has been (highly likely) assigned with to an incorrect section (see Section \ref{level3}). Hence, the input is a short text together with a label assigned by the user. The system either issues a warning about the assignment being likely incorrect or abstains. 

We transform the dataset by creating text-label pairs between each sentence and all the labels. Since there are 1,440 sentences in the test set and 6 unique labels, there are 8,640 such pairs. For each pair, we retrieve the probability distribution the model assigns over the possible labels. 
If the value for a given pair is below a static threshold (we experiment with 0.05, 0.1 and 0.2), the system issues a warning. It is crucial to minimize the number of false positives (i.e., issuing a warning when the user-assigned label is in fact correct). It is comparatively less important to treat false negatives (i.e. missing out on an incorrect assignment). Since the user is still in the process of learning, abstaining in case of a mistake is preferable to providing the user erroneous (confusing) feedback.

The results for the three thresholds are reported in Table \ref{tab:errors}. The columns correspond to whether the warning should be raised, whereas the rows correspond to whether the model would raise the warning or abstain. The diagonal (cells shaded in green) reports the number of pairs for which the model behaves as desired. The cells outside of the diagonal (shaded in red) report the disagreements.

\begin{table}
    \caption{Statistics about warning and abstentions for section assignments, for thresholds 0.05, 0.1 and 0.2.}
{\footnotesize
    \begin{tabular}[t]{lrr}
        &Warn & Abstain \\ 
        Warn & \cellcolor{green!25}5952& \cellcolor{red!25}122 \\
        Abstain &\cellcolor{red!25}1248 & \cellcolor{green!25}1318\\ 
    \end{tabular}
    \hfill
    \begin{tabular}[t]{lrr}
        &Warn & Abstain \\ 
        Warn & \cellcolor{green!25}6282& \cellcolor{red!25}169 \\
        Abstain &\cellcolor{red!25}918 & \cellcolor{green!25}1271\\ 
    \end{tabular}
    \hfill
    \begin{tabular}[t]{lrr}
        &Warn & Abstain \\ 
        Warn & \cellcolor{green!25}6549& \cellcolor{red!25}256 \\
        Abstain &\cellcolor{red!25}651 & \cellcolor{green!25}1184\\ 
    \end{tabular}}
    \label{tab:errors}
\end{table}

\subsection{Testing H2: Categorizing Key Argument Elements Automatically}
When interacting with the Advanced and Expert users (NIH Levels 4 and 5) CABINET automatically categorizes the key argument elements identified in the text by the user (see Section \ref{level4}). This is a straightforward sentence classification task. For this component, a certain amount of error is tolerable since (a) the user's proficiency is relatively high and (b) the user is actively involved in selecting the sentences. Hence, they are in a good position to verify and potentially correct and confirm the system's category assignment. 

To evaluate the ability of automatically categorizing argument elements, we compare predictions of the trained RoBERTa model to the baseline. As shown in  Table \ref{roberta_results}, it appears the performance differs considerably across types. Facts and Issue argument element types are identified more reliably than Holding or Rule. Table \ref{roberta_results} shows a confusion matrix over which classes are frequently confused with other classes. The predicted labels are shown in the rows, and the true labels in the columns. For example, we can see that holdings are frequently confused with reasoning, which may be due to the small size of the classes or the classes having low ``semantic homogeneity'', compare \cite{westermann2021}.

\begin{table}[t]
        \centering
        \setlength{\tabcolsep}{4.5pt}
        \caption{Performance (left) and confusion matrix (right) for class predictions.}
        \begin{tabular}{l|rrr|rrrrr|cccccc}
                           & \multicolumn{3}{c|}{Baseline} & \multicolumn{3}{c}{RoBERTa} & & & \multirow{3}{*}[-5pt]{\rotatebox[origin=b]{90}{Facts}} & \multirow{3}{*}[-5pt]{\rotatebox[origin=c]{90}{Issue}} & \multirow{3}{*}[-.5pt]{\rotatebox[origin=c]{90}{Holding}} & \multirow{3}{*}{\rotatebox[origin=c]{90}{Proc. H.}} & \multirow{3}{*}[-1pt]{\rotatebox[origin=c]{90}{Reason}} & \multirow{3}{*}[-5pt]{\rotatebox[origin=c]{90}{Rule}} \\
        Argument Type      & P & R & $F_1$ & $P$ & $R$ & $F_1$ & &   \\
        \cline{1-7}
        Facts              & .46  & .47  & .47  & .90 & .80 & .85 & &   \\
        \cline{9-15}
        Issue              & .05  & .05  & .05  & .96 & .91 & .93 & & Facts & \cellcolor{blue!100}\textcolor{white}{524} & \cellcolor{blue!0}0 & \cellcolor{blue!2}9 & \cellcolor{blue!3}14 & \cellcolor{blue!6}31 & \cellcolor{blue!1}3  \\

        Holding            & .13  & .17  & .15  & .42 & .53 & .47 & & Issue & \cellcolor{blue!0}2 & \cellcolor{blue!23}122 & \cellcolor{blue!0}0 & \cellcolor{blue!0}0 & \cellcolor{blue!0}0 & \cellcolor{blue!1}3  \\

        Procedural History & .06  & .06  & .06  & .66 & .81 & .73 & & Holding & \cellcolor{blue!5}25 & \cellcolor{blue!1}4 & \cellcolor{blue!14}72 & \cellcolor{blue!1}4 & \cellcolor{blue!9}46 & \cellcolor{blue!4}21  \\
        Reasoning          & .23  & .17  & .20  & .56 & .67 & .61 & & Proc. H. & \cellcolor{blue!7}35 & \cellcolor{blue!1}3 & \cellcolor{blue!1}6 & \cellcolor{blue!18}95 & \cellcolor{blue!1}3 & \cellcolor{blue!0}1  \\
        Rule               & .06  & .07  & .07  & .65 & .48 & .55 & & Reason & \cellcolor{blue!11}59 & \cellcolor{blue!1}4 & \cellcolor{blue!7}36 & \cellcolor{blue!1}5 & \cellcolor{blue!35}181 & \cellcolor{blue!7}38  \\
        \cline{1-7}
        Weighted Avg       & .28  & .27  & .27  & .76     & .73      & .74  & & Rule & \cellcolor{blue!2}8 & \cellcolor{blue!0}1 & \cellcolor{blue!2}13 & \cellcolor{blue!0}0 & \cellcolor{blue!2}11 & \cellcolor{blue!12}61  \\
        \end{tabular}

        \label{roberta_results}
\end{table}

\section{Discussion and Future Work}
The results of the experiment evaluating H1 show that the false positive rate (see Table \ref{tab:errors}) varies between 2.0\% and 3.8\%, depending on the  threshold. This rate appears to be acceptable given the envisioned use case and user's proficiency level (Intermediate - NIH Level 3). The false negatives rate (i.e., the system abstains when a warning should have been raised) varies between 48.6\% and 35.5\%. While such a rate is  high, we argue that in case of an isolated error due to factors such as fatigue, stress, or lack of attention, the missed warning is tolerable. If a learner has a systematic misconception, the user errors will repeat and the system will likely detect a larger portion of those. Hence, the learner will receive clear and timely feedback triggering further learning. 

Evaluation of H2 shows promising performance of the fine-tuned RoBERTa model, although the performance is far from perfect. This is acceptable since this component supports a user at Advanced or Expert proficiency level (NIH Levels 4 and 5). Therefore the potential to confuse a user by an incorrect prediction is relatively low. Since the user actively selects the argument element and is immediately presented with a prediction they are in a comfortable position to perform a correction. We argue that it is far more efficient to inspect automatic predictions and make corrections when needed (in about 25\% of predictions), compared to categorizing the key argument elements  manually. 

While the experimental results confirm our working hypotheses, there are several important limitations to the presented study. 
\textit{First}, the design of the experiments only takes into account passages of text that have been selected for inclusion in the case briefs by legal experts. However, the user may occasionally make mistakes in their selections. This is particularly true for users at the Intermediary proficiency level (NIH level 3).
\textit{Second}, we do not address the challenge of assessing the current level of proficiency of the user. 
\textit{Third}, the functionality of the system at the Fundamental Awareness (NIH Level 1) proficiency level requires a limited but highly curated dataset of annotated cases with detailed feedback addressing common misconceptions---a resource we have not yet created. 
\textit{Fourth}, we did not conduct a feasibility study of the highlighting functionality at the Expert proficiency level (NIH Level 5). 
\textit{Fifth}, and most importantly, we have not conducted a pilot user study to tentatively gauge the expected improvements in learning outcomes. We plan to address these limitations in future work.

\section{Conclusion}
We proposed an adaptive environment to support case law analysis based on a novel cognitive computing framework that  matches various ML capabilities to the proficiency of a user. We have shown how the environment could (i) support a learner in mastering the skill of identifying key argument elements in a court opinion, and (ii)  support a professional in performing the same task more efficiently. We have demonstrated that it is possible to detect if a sentence is placed in an incorrect section in case brief (H1), and to predict the actual argument element type of a case brief sentence (H2) with a reliability sufficient for the envisioned use case based on the proficiency level of a user. Hence, we have taken the initial steps in establishing the feasibility of the proposed system.

\noindent\rule{0pt}{4ex}\textbf{Acknowledgements} 
Hannes Westermann and Karim Benyehklef acknowledge the generous support from the 
Cyberjustice Laboratory,  LexUM Chair on Legal Information, and  Autonomy through Cyberjustice Technologies project. Figure \ref{fig:framework} has been designed using resources from \url{Flaticon.com}.

\end{document}